# SELF LEARNING ROBOT USING REAL-TIME NEURAL NETWORKS

Chirag Gupta[1], Chikita Nangia[2], Chetan Kumar[3]

[1]Department of Electronics and Communication Engineering, Thapar University, Patiala, India
[2]Department of Electronics and Communication Engineering, Thapar University, Patiala, India
[3]Department of Electronics and Communication Engineering, Thapar University, Patiala, India

**Abstract**

*With the advancements in high volume, low precision Computational technology and applied research on cognitive Artificially Intelligent heuristic systems, Machine Learning solutions through Neural Networks with real-time learning has seen an immense interest in the research community as well the Industry. This Paper involves Research, Development and Analysis of a Neural Network implemented on a robot with an arm through which evolves to learn to walk in a straight line or as required. The Neural Network learns using the Algorithms of Gradient Descent and Backpropagation. Both the implementation and training of the Neural Network is done locally on the robot on a Raspberry Pi 3 so that its learning process is completely independent. The Neural Network is first tested on a Simulator developed on MATLAB and then implemented on Raspberry Pi 3. Data at each generation of the evolving network is stored, and analysis both mathematical and graphical is done on the data. Impact of factors like Learning Rate and Error Tolerance on the learning process and final output is analyzed.*

--------------------------------------------------------------------***--------------------------------------------------------------------

## 1. INTRODUCTION

With the advancement in portable low power processing solutions in recent years, real-time Neural Networks [1] implemented locally on Embedded and Robotic Systems have become more practical as well as an interesting area of research. Robotic systems such as those for defence, UAVs or space probes going into uncharted extraterrestrial territory can benefit greatly from heuristic self learning models implemented locally as external communication and interference might be difficult or infeasible in such systems. Such tasks like learning to navigate through complex terrain or dealing with new problems not encountered before are extremely difficult to implement using traditional Logic based programming techniques but can be effectively dealt with machine learning through neural networks.

### 1.1 Neural Network

A Neural Network (NN) or an Artificial Neural Network (ANN) is a computational Information processing model based on the biological neural network. Neural network works in the same ways as our brain does by learning from its experience. The network has large number of inter-connected processing neurons. A neural network has one input layer, one output layer and can have multiple hidden layers. In Real-time neural networks, the neurons take inputs from other sources in its environment and combine them by performing some non-linear operation to get the final output. Information is processed collectively, as the all the neurons work in a parallel way. Artificial neurons communicate by sending signals to each other through a large number of weighted connections. As the network finds out how to solve the problem by itself, its operation can be unpredictable. The processing ability of the network is stored in the strengths or weights of connection, obtained by a process of adaptation to, or learning from, a set of training patterns.

### 1.2 Neural Network Algorithms

During the learning process the neural network structure is altered by increasing or decreasing the strength of connections. More relevant information will have stronger connections and less relevant information will have weak connection strength. This strengthening and weakening of the connections enables the network to learn.

We start to train the neural network by choosing a random parameter. Then, sequence of parameters is generated, so that the error is reduced at each generation of the algorithm. The training algorithm stops when the output is within our desired tolerance levels. The different training algorithms employed are:

Gradient descent is the simplest training algorithm for optimization. It requires information from the gradient vector, and therefore it is the first order method. Gradient descent is known as steepest descent. Gradient descent method is used to determine the weights so as to move in the direction of minimum error [2].

The main purpose is to optimize the cost function. Gradient Descent is used to minimize the cost function. Gradient descent is used to find local minima of a function. We start with an initial arbitrary guess of the solution and take the gradient of function at that point and step the solution in the opposite direction of the gradient and repeat this process. This algorithm converges to the point where the gradient is zero.

Backpropagation algorithm trains a given feed-forward neural network for a given set of input patterns. The training of a neural network involves three stages: feedforward of the input training pattern, the calculation and Backpropagation of the associated error and the adjustment of weights. During feedforward, each input neuron receives an input signal and then sends it to each hidden layer which computes its output using an activation function and then sends it to each output unit. Each output computes its activation. During training, each output compares its activation with the desired output to calculate the associated error. And this error is propagated back to the previous layer and updates the weights and biases of connection between output and hidden layer. And in the same way error is calculated for each hidden unit and this error is used to update the weights of connections between input and hidden layer.

## 2. DESIGN

The objective of this paper is to develop a real-time neural network (training data generated in real time) implemented locally (on the robot itself) on a robot with a single arm through which it evolves to learn to walk in a straight line or as required. Gradient Descent and Backpropagation Algorithms and Network Structures are tested and analyzed for their performance in a special as well as a general setting. [3]

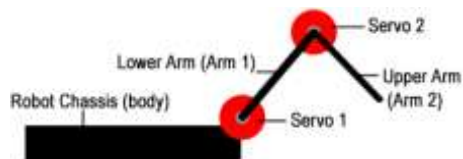

**Fig 1:** Physical Structure of the Robot

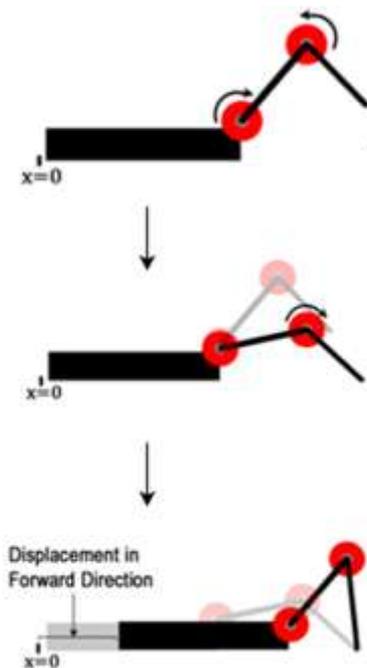

**Fig 2:** Required (Objective) Motion of the Robot

### 2.1 Hardware Specifications

**Table 1:** Hardware Specification

| Specification | Component | Reasons |
|---|---|---|
| Processing and Control | Raspberry Pi 3 Model B | 1.2 GHz Quad Core ARM CPU capable of running neural networks<br>1 GB RAM require for storing real time variables like weights, biases etc.<br>Full Desktop class Linux OS<br>Low Power Consumption: 1.5 W (average); 6.7 W (max) |
| Accelerometer and Gyroscope | MPU6050 | 3 axis MEMS Accelerometer<br>3 axis MEMS Gyroscope<br>Easy I2C Interfacing<br>Compatible with 3.3V-5.0V voltage level<br>Inexpensive |
| Servos (1 and 2) | High Torque Plastic Gear Servo Motor | High Torque needed to move the Arm using friction.<br>Torque: 1.8 kg-cm<br>Operating Speed: 0.10 sec/60 degree<br>Rotational Range: 180°<br>Operating Voltage: 4.8V |
| Power | Battery for Electronics and Servos | Power Bank: 5V, 2.1A USB Power Source for Raspberry Pi Board<br>9V Battery for Servos and Other Electronics |
| Chassis | Plastic and Metal off-the-shelf robotic chassis components | Both Plastic and Metal Components are considered as per availability and price |

## 2.2 Software Specifications

**Table 2:** Software Specification

| Programming Language | Java 8 on IntelliJ IDEA IDE | Simple but Advanced Object Oriented features Extensive Native Libraries for a large number of functions Libraries and Frameworks available to use with Raspberry Pi Easy effective GUI development with Java Swing and Java Graphics |
|---|---|---|
| Neural Network Data Structure | Feedforward Neural Network which is an object of the Network Class | 3 Layers: 1 Input layer, 1 Hidden layer, 1 Output Layer No of input neurons: 1 No of hidden layer neurons: Variable No of output neurons: 2 Activation Function: Sigmoid Function |
| Training Algorithm | Gradient Descent and Backpropagation | Gradient Descent to achieve global minima of error distribution. Backpropagation to change the neural attributes like weights |
| Simulator and Data Analysis | MATLAB | Easy and Dynamic Plot and Erase Functions Extensive Libraries of general and higher mathematical functions 3 dimensional and subplotting features |

## 2.3 Process

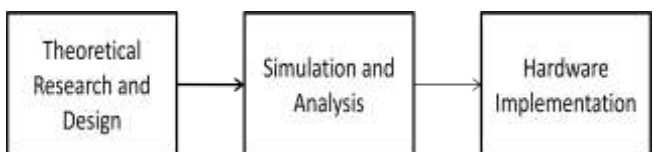

**Fig 3:** Stage Process Methodology

The first stage includes Research on the concepts, mathematics, published literature and previous work as well as any current open source projects if any on that particular topic. Then Mathematical formulation and other constants are devised in accordance with that research to provide a skeleton to be used while programming the network. Finally it is designed both on paper as well as a 3d model is designed in 3DS Max and rendered with Mental Ray to provide photorealistic visualization of the physical model we are trying to achieve.

The second stage includes the Simulation of the Network for Quantitative and Qualitative analysis of the network on software. First the theoretical and mathematical network model developed in the first stage is programmed in the Java Language and test run. The code written is highly modular so that same code can be reused for different implementations. The network code is then tested and visualized on a simulator. We have developed a custom 2D simulator for one armed robot in MATLAB for experiments done in this report. Data at each generation is saved and the complete data of the learning process is then processed to produce tables, graphs and visualization to analyze error at each generation, learning process, effectiveness of various parameters etc analysis. The model and code is modified till the results obtained are satisfactory. Results up to this stage are presented in this report.

Third Stage is the Hardware implementation of the neural network model developed in previous stages. It is implemented in a Raspberry Pi 3 board which processes the code and sends the output angle signal to the actuators that is the servo motors of the robot arm. It goes through successive generations till the robot learns to walk in straight line or any other motion as required.

At start the Learning Rate and the Error Tolerances are set by the user and the Generation is set to 1. A random input is generated for the input neuron and the feedforward operation is done on the neural network to obtain the two normalized outputs. They are then denormalized and the two angles in degrees for the two servo motors are obtained. They are then fed to the corresponding motors which move according to the angles, consequently moving the robot. The final position and orientation is calculated using the accelerometer and gyroscope of the MPU6050 [4]. If those values are as desired or within the tolerance levels, the network is deemed trained and the same is saved on secondary memory before the termination of the program. However if the error is more than the tolerance level, the Generation is incremented by 1, results are normalized and fed into the gradient descent algorithm which calculates the direction and delta values of maximum decrease of error, and the attributes (weights, biases) are then updated using the backpropagation algorithm, and then the feedforward is done again.

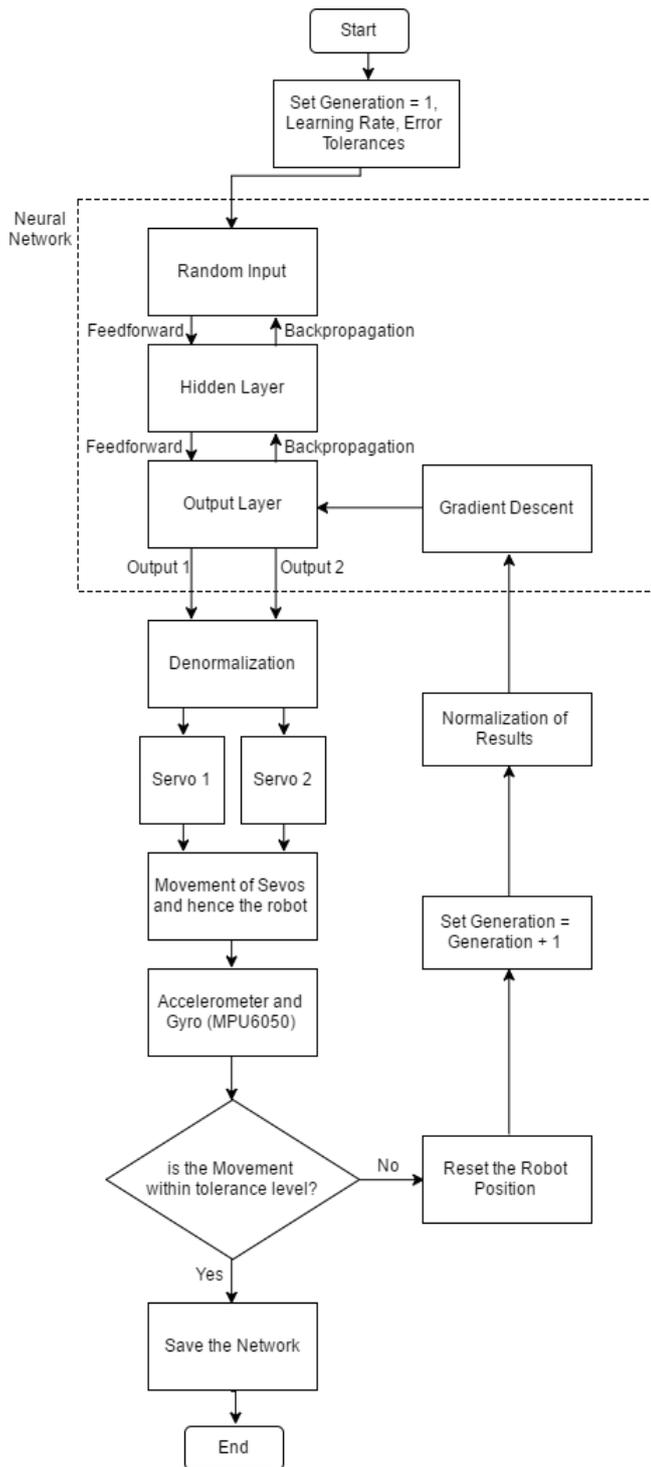

**Fig 4:** Process Flow Chart

## 2.4 Functional Description

The neural network consists of 3 layers each containing a variable number of neurons. As the robot has to learn to walk in a desired manner irrespective of the input, the network doesn't need any particular input. Consequently the input layer consists of only one neuron having an arbitrary value. The Hidden layer can have any integral number of neurons based on the complexity of the problem. The size of the hidden layer i.e. the number of hidden neurons is provided at the initialization of the program. The output consists of 2 neurons representing the angles for the servo motors to move.

Each Neuron except the input neuron has a Summer ($\sum$), bias and an Activation Function. The connections between any two neurons are weighted that is it carries a weight which is a decimal value between -1 and 1. The summer sums the weighted inputs (i.e. $\sum w_i * x_i$) directed towards it that is coming into the neuron. It is then passed through an activation function which gives an output between 0 and 1.

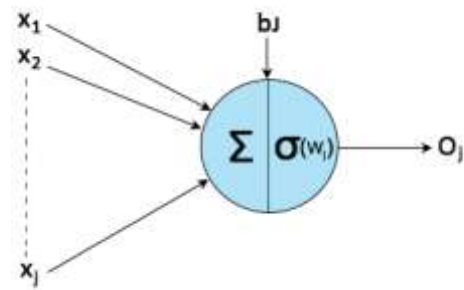

**Fig 5:** Neuron Model

First the summer sums the weighted inputs,

$$W_j = \sum_{k=1}^{i} w_{kj} * x_k + b_j$$

Where
$x_k$ is the input to the $j^{th}$ neuron of current layer from $k^{th}$ neuron of previous layer,
$w_{kj}$ is the weight from neuron k to neuron j,
$b_j$ is the bias of $j^{th}$ neuron,
$W_j$ is the output of the Summer.

Now Wj is passed through the sigmoid activation function to generate the neuron output,

$$O_j = \sigma(W_j)$$

After getting the final result of the feedforward network in the output layer, the two outputs are the denormalized to get the angles in degrees. These are then fed to the robot in the simulator which simulates the motion by rotating the servos according to the angles and then the error is calculated using gradient descent.

The activation function decides if a neuron fires (ON i.e. output = 1) or not (ON i.e. output = 0). If the input to the activation function exceeds some threshold value, the neuron fires. This threshold value is taken to be 0. The activation function used is Binary Sigmoid which is,

$$\sigma(x) = \frac{1}{1 + e^{-x}}$$

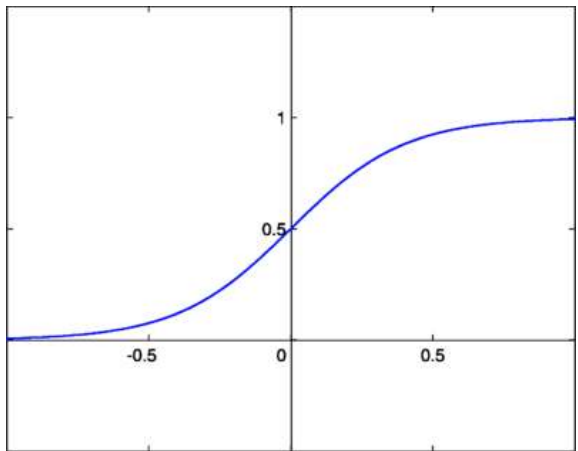

**Fig 6:** Sigmoid Function

The role of activation function is to make our neural networks non-linear. Sigmoid function is suitable for this job as it is both non-linear and differentiable at all points. It gives HIGH (1) output for large inputs and LOW (0) output for small inputs but for inputs closer to 0 it gives a range of fractions between 0 and 1.

The required outputs of the feed forward neural network are the two angles to be fed to the two servo motors. The angles range from 0° to +180°, however the neural networks has outputs of each neuron ranging from 0 to 1 which is the range for the sigmoid activation function. To solve this problem, angles in degrees are normalized to a decimal value between 0 and 1 and conversely the output of the network is denormalized to angles in degrees to feed into the servo motors. The normalization and denormalization are linear functions as given below.

Normalization:

$x = N(y) = (y/180.0)$    $0 \leq x \leq 1.0$
$0 \leq y \leq 180.0$

Denormalization:

$y = N(x) = (180 * x)$    $0 \leq y \leq 180.0$
$0 \leq x \leq 1.0$

## 3. SIMULATION AND ANALYSIS

The Error is calculated by estimated required values of the Angles.

### 3.1 Case 1

Number of Hidden Neurons: 2
Learning Rate = 0.8
Error Tolerance = 1°

The network (robot) takes 148 iterations (generation) to train (learn):

**Table 3:** Generation Table of Evolving Neural Network (Case 1)

| Generation | Servo 1 (°) | Servo 2 (°) | Error 1 (°) | Error 2 (°) |
|---|---|---|---|---|
| 1 | -73.44 | -39.024 | 163.44 | 159.024 |
| 2 | -69.48 | -35.064 | 159.48 | 155.064 |
| 3 | -64.656 | -30.6 | 154.656 | 150.6 |
| 4 | -61.056 | -27 | 151.056 | 147 |
| 5 | -57.96 | -23.688 | 147.96 | 143.688 |
| 10 | -40.824 | -6.984 | 130.824 | 126.984 |
| 15 | -28.008 | 7.92 | 118.008 | 112.08 |
| 20 | -13.032 | 21.6 | 103.032 | 98.4 |
| 30 | 10.008 | 41.832 | 79.992 | 78.168 |
| 50 | 42.48 | 73.656 | 47.52 | 46.344 |
| 75 | 64.152 | 93.744 | 25.848 | 26.256 |
| 100 | 71.208 | 99.144 | 18.792 | 20.856 |
| 148 | 90.216 | 120.384 | 0.216 | -0.384 |

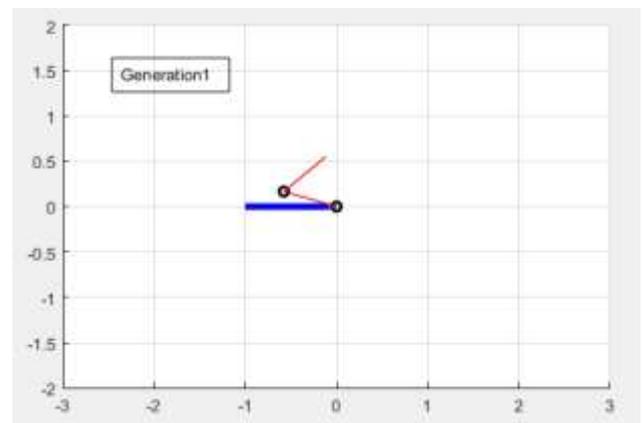

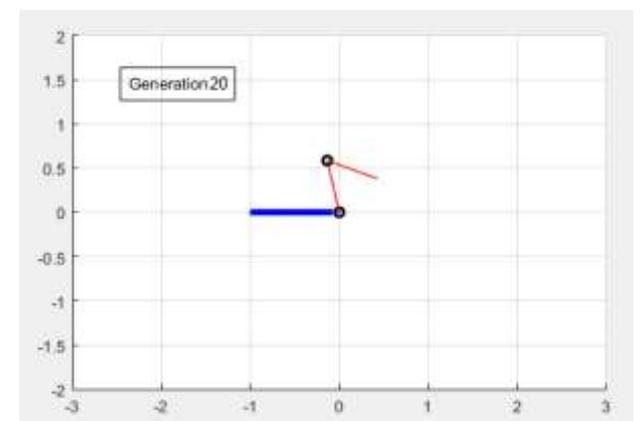

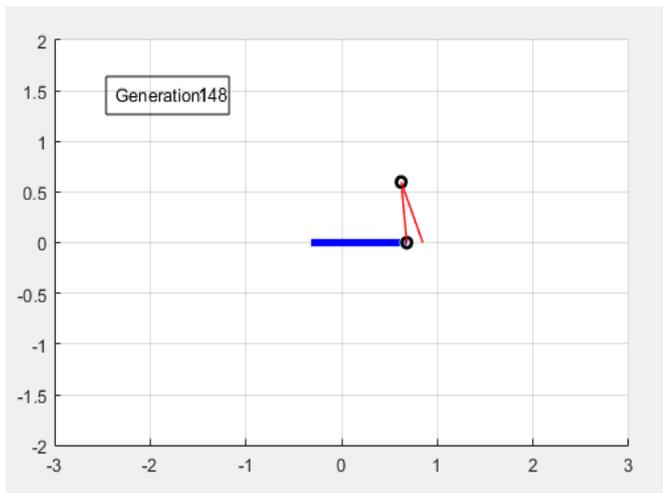

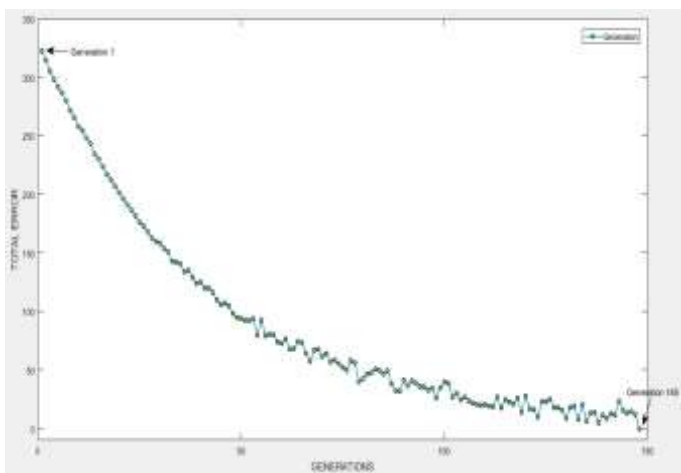

**Fig 7:** Evolution of Neural Network

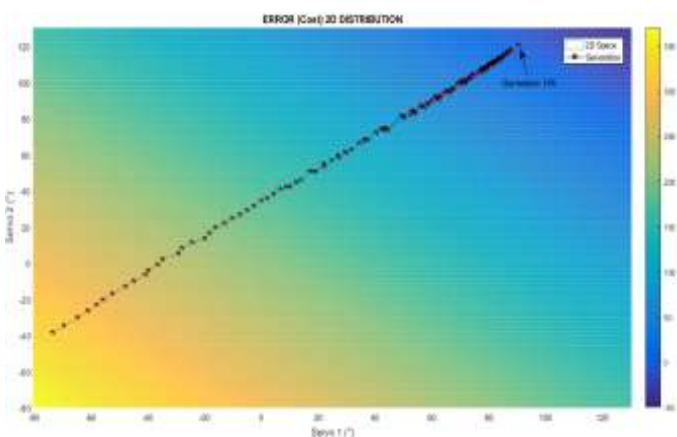

**Fig 8:** Generation Error Scatter Plot

## 3.2 Case 2

Number of Hidden Neurons: 2
Learning Rate = 0.5
Error Tolerance = 5°

The network (robot) takes 156 iterations (generation) to train (learn):

**Table 4:** Generation Table of Evolving Neural Network (Case 2)

| Generation | Servo 1 (°) | Servo 2 (°) | Error 1 (°) | Error 2 (°) |
|---|---|---|---|---|
| 1 | -60.264 | 27 | 150.264 | 93 |
| 2 | -57.168 | 29.232 | 147.168 | 90.768 |
| 3 | -52.632 | 32.976 | 142.632 | 87.024 |
| 4 | -49.176 | 35.496 | 139.176 | 84.504 |
| 5 | -51.912 | 32.256 | 141.912 | 87.744 |
| 10 | -43.272 | 37.224 | 133.272 | 82.776 |
| 15 | -35.424 | 41.616 | 125.424 | 78.384 |
| 20 | -23.904 | 49.248 | 113.904 | 70.752 |
| 30 | -6.984 | 59.544 | 96.984 | 60.456 |
| 50 | 28.584 | 82.152 | 61.416 | 37.848 |
| 75 | 48.6 | 93.384 | 41.4 | 26.616 |
| 100 | 52.992 | 94.608 | 37.008 | 25.392 |
| 156 | 86.616 | 117.288 | 3.384 | 2.712 |

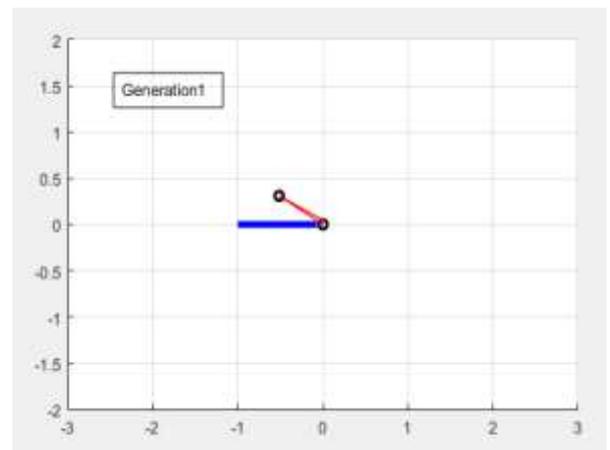

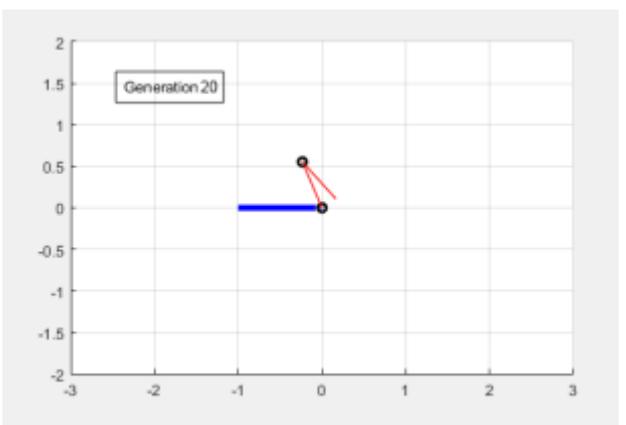

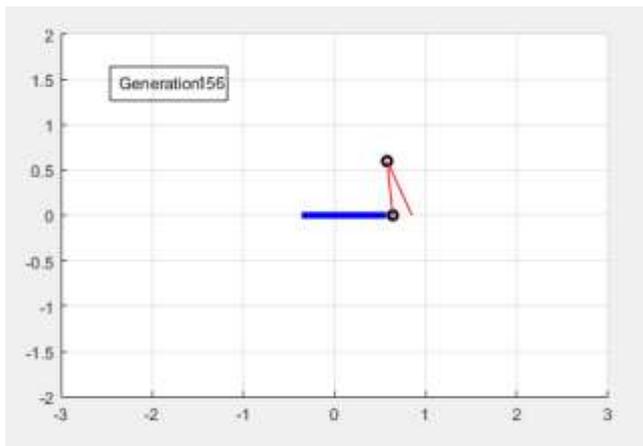

**Fig 9:** Evolution of Neural Network

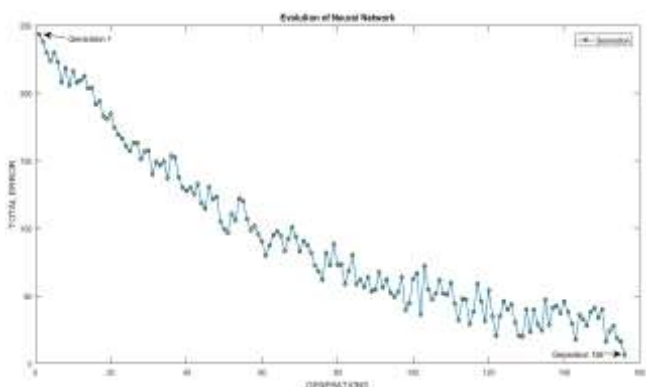

**Fig 10:** Generation Error Scatter Plot

## 4. IMPLEMENTATION

Neural Network was implemented on the robot using Raspberry Pi 3 as the processor and the platform for the robot, MPU6050 Accelerometer/Gyro to track the robot's movement inorder to calculate the error to be fed into the backpropagation algorithm. The chassis and the full structure of the robot was built using both off the shelf components and also by fabricating custom pieces (e.g. steel rod pair) wherever necessary using machining, fitting etc. in the Workshop. The robot structure has resulted in a strong and fine piece of craftsmanship built to withstand all the strong jerks and moment of inertias it is subject to due to its long arm and servos during training. Therefore to keep its structural integrity mechanical nut-bolts (over 30) are used extensively wherever possible, adhesive solutions are employed elsewhere. Again, for the robot to absorb strong moments its center of gravity has been kept low (towards the ground) and in the middle slightly towards its back. Extra strap on weight has been added at various places to achieve the same.

The Neural Network software is a modified one which was used during simulation. It is written JAVA running on a Raspbian (Linux) JVM on the Pi. A few GUI elements (e.g. learning progress bar has been added). However everything else including the movement of servos, accelerometer data, indicator leds are programmed in python in a separate program. This decision was taken due to the extensive libraries available and relative development ease of python partly due to python being the 'default' language of Raspberry Pi [5]. The two programs communicate signals and data seamlessly using a simple custom protocol involving reading and writing text files.

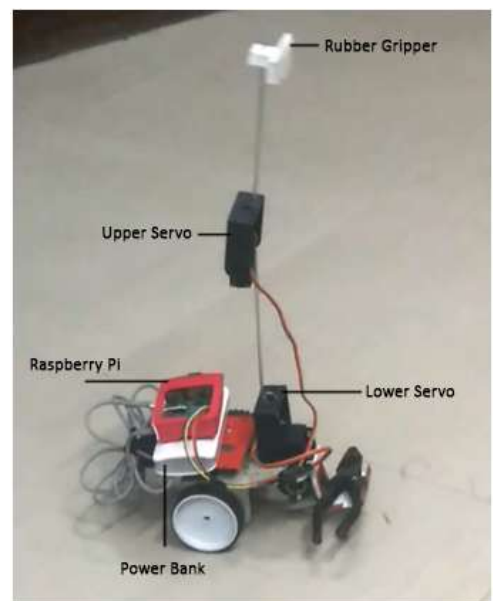

**Fig 11:** Robot Structure

Number of Hidden Neurons: 20
Learning Rate = 0.3

Generation 1:

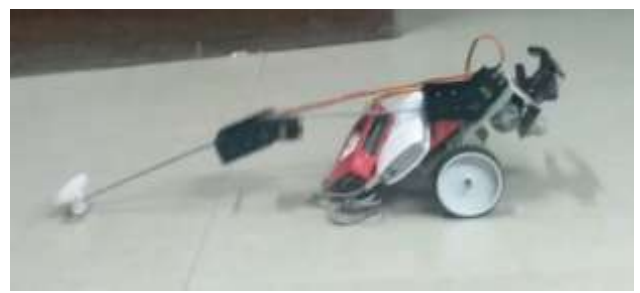

Generation 20:

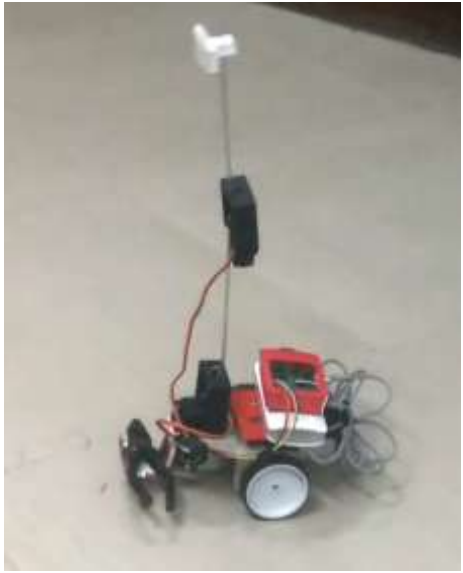

Generation 56:

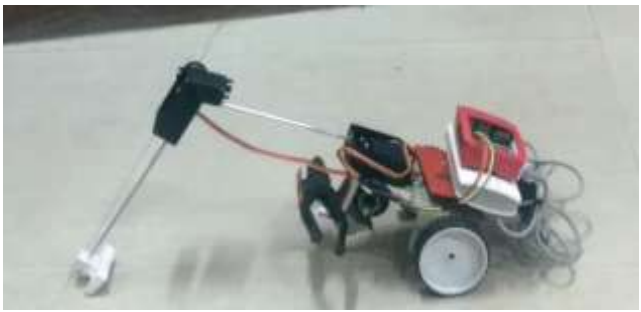

## 5. CONCLUSION

The Neural Network worked flawlessly almost every time resulting in the robot learning the desired pattern to move its limb so as to crawl in a straight line. Therefore the most important objective of a TD (technology demonstration) which implemented neural network on a standalone low computational intense platform and training it in real-time using real-time experience with no external control requirement has been successfully achieved.

The training data is mostly consistent with the theory as well as that achieved during simulation. The learning process is affected mainly due to the following two choices taken before the learning is begun:

**(i) Hidden Layer Size**
The speed of the learning process i.e. fewer number of generations is seen to be directly proportional to the hidden layer size or the number of neurons in the hidden layer. A large hidden layer allows for more abstraction to happen and this aspect world be imperative for more complex tasks such as complex gestures.

However the increase in learning efficiency with hidden layer size is not linear but decreases exponentially with capping around 25 neurons. This decrease can be explained by the relatively simple task which is the objective here, however further research is required o verify the same.

**(ii) Learning Rate**
The Learning rate is directly responsible for the speed of the learning process. A large learning rate (greater than 0.4) increases the learning process but may miss the optimization point. A smaller learning rate learns slowly but is better of achieving the global minimum error. Therefore a dynamic approach is best keeping learning rate larger and first and then decreasing it gradually. The magnitudes are chosen by analyzing the specific objective and refining them by trial.